%% file: main.tex
\documentclass[10pt,twocolumn,letterpaper]{article}

\usepackage{cvpr}      

\usepackage{mathtools}
\usepackage{graphicx}
\usepackage{subcaption} 
\usepackage{caption}
\usepackage{booktabs}
\usepackage{makecell}

\newcommand{\modelname}{BrepARG}

\input{preamble}
\definecolor{cvprblue}{rgb}{0.21,0.49,0.74}
\usepackage[pagebackref,breaklinks,colorlinks,allcolors=cvprblue]{hyperref}

\def\paperID{} 
\def\confName{CVPR}
\def\confYear{2026}

\title{AutoRegressive Generation with B-rep Holistic Token Sequence Representation}
\author{
Jiahao Li\textsuperscript{1} \quad 
Yunpeng Bai\textsuperscript{2} \quad 
Yongkang Dai\textsuperscript{1} \quad 
Hao Guo\textsuperscript{1} \quad 
Hongping Gan\textsuperscript{1} \\
Yilei Shi\textsuperscript{1}\thanks{Corresponding author.} \\
\textsuperscript{1}Northwestern Polytechnical University \quad 
\textsuperscript{2}National University of Singapore \\
{\tt\small \{lijiahao142857,daiyongkang,guoh0215\}@mail.nwpu.edu.cn} \\
{\tt\small \{ganhongping,yilei\_shi\}@nwpu.edu.cn,\ bai\_yunpeng99@u.nus.edu}
}

\begin{document}
\maketitle
\begin{abstract}
Previous representation and generation approaches for the B-rep relied on graph-based representations that disentangle geometric and topological features through decoupled computational pipelines, thereby precluding the application of sequence-based generative frameworks, such as transformer architectures that have demonstrated remarkable performance. In this paper, we propose \textit{\modelname}, the first attempt to encode B-rep's geometry and topology into a holistic token sequence representation, enabling sequence-based B-rep generation with an autoregressive architecture. Specifically, \textit{\modelname} encodes B-rep into 3 types of tokens: geometry and position tokens representing geometric features, and face index tokens representing topology. Then the holistic token sequence is constructed hierarchically, starting with constructing the geometry blocks (i.e., faces and edges) using the above tokens, followed by geometry block sequencing. Finally, we assemble the holistic sequence representation for the entire B-rep. We also construct a transformer-based autoregressive model that learns the distribution over holistic token sequences via next-token prediction, using a multi-layer decoder-only architecture with causal masking. Experiments demonstrate that \textit{\modelname} achieves state-of-the-art (SOTA) performance. \textit{\modelname} validates the feasibility of representing B-rep as holistic token sequences, opening new directions for B-rep generation. The source code is available at \url{https://github.com/123qiang06/BrepARG}.

\end{abstract}
\begin{figure}[!t]
    \centering
    \includegraphics[width=\columnwidth]{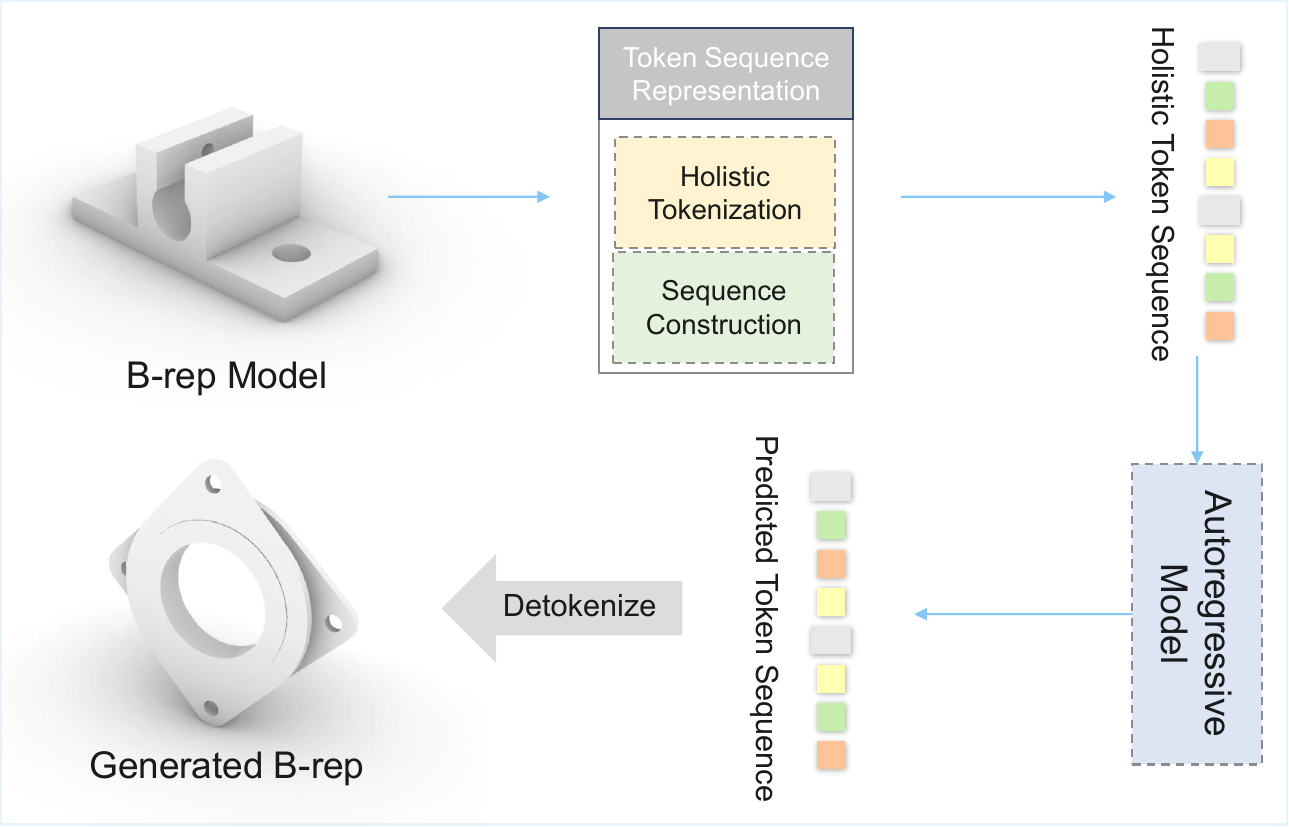}
    \caption{\textit{\modelname} is the first approach to encode the geometry and topology of a B-rep into a holistic token sequence, enabling efficient modeling and generation via an autoregressive model.}
    \label{fig:BrepARG}
\end{figure}
\section{Introduction}

Boundary Representation (B-rep)~\cite{brep} is a fundamental paradigm in Computer-Aided Design (CAD) for representing solid models. Recent progress in B-rep data generation~\cite{brepgen,dtgbrepgen,solidgen,brepgiff} has advanced the automation and learning of CAD modeling processes. However, B-rep structures are inherently complex: they combine heterogeneous geometric primitives (parametric faces, edges, and vertices) within hierarchical topological dependencies. This heterogeneity makes it difficult to encode B-rep data in a direct, holistic form suitable for deep generative modeling.

Existing methods fall short in representing geometric and topological features holistically. They generally rely on stage-wise learning paradigms that separately model geometry and topology~\cite{brepgen,dtgbrepgen,solidgen}, or employ multi-component architectures for different elements~\cite{brepgiff}. Such designs lead to fragmented representations and increased model complexity. Moreover, graph-based formulations~\cite{brepgen,brepgiff} constrain the use of efficient transformer-based architectures that require sequential inputs, while multi-stage approaches~\cite{dtgbrepgen,solidgen} limit the model’s ability to capture the full heterogeneity and interdependence inherent in B-rep.

\begin{figure*}[!t]
    \centering
    \includegraphics[width=\textwidth]{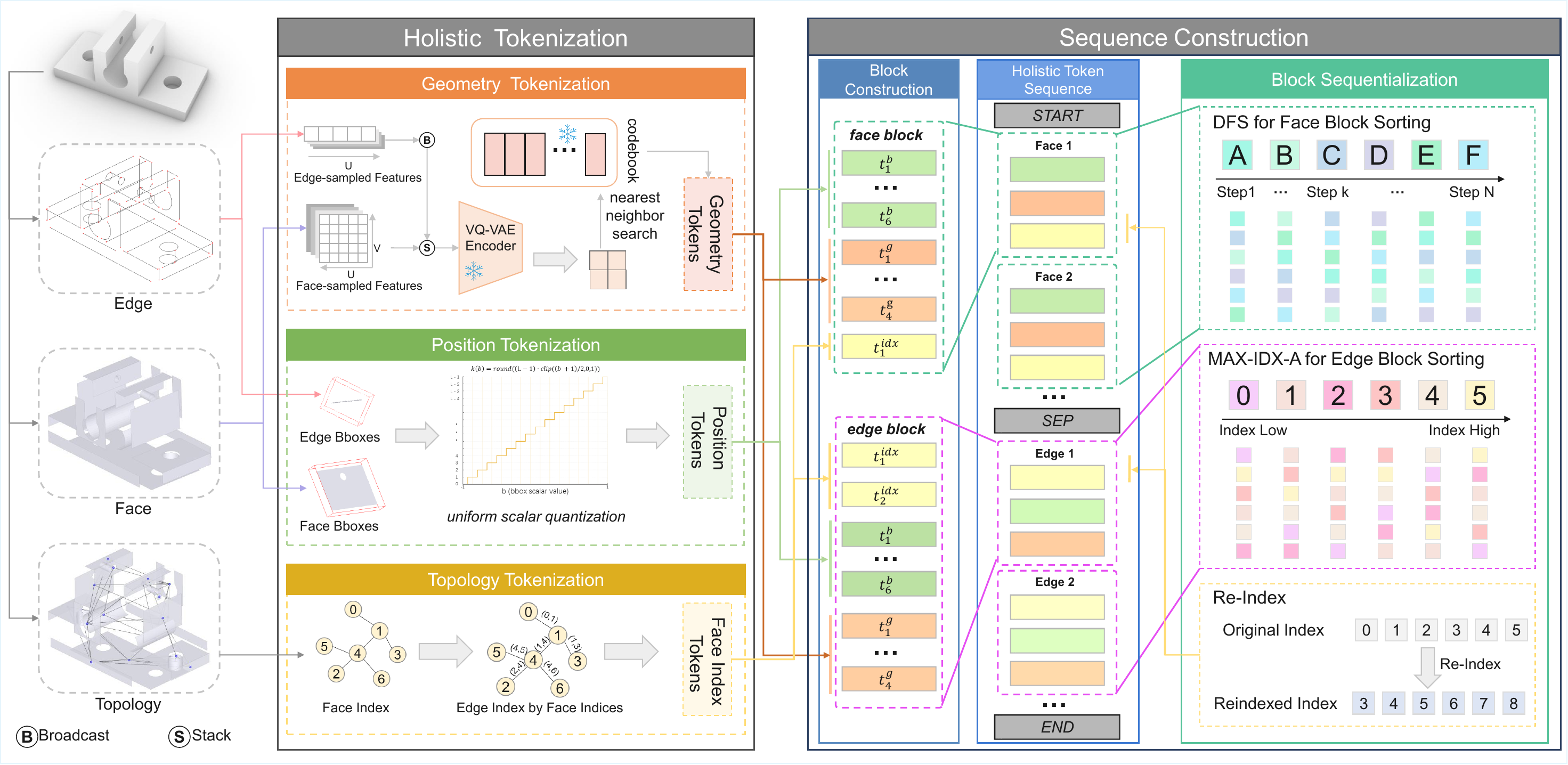}
    \caption{Our Holistic Token Sequence Representation comprises two parts: 1) \textbf{Holistic Tokenization}: \textit{Geometry Tokens} are constructed  by mapping VQ-VAE encoder latents to the codebook. \textit{Position Tokens} are constructed by applying uniform scalar quantization to bounding boxes (Bboxes). \textit{Face Index Tokens}, used for conveying topology information, are constructed by assigning an index to each face. 2) \textbf{Sequence Construction}: We first assemble face and edge blocks using the three token types above. These blocks are then ordered independently using DFS and the MAX-IDX-A algorithm. Finally, the ordered face and edge block sequences are concatenated with a separator and wrapped with start/end-of-sequence markers, producing the holistic token sequence representation of the entire B-rep model.}
    \label{fig:sequence}
\end{figure*}

To holistically represent B-rep, in this work, we propose a \textbf{B-rep} \textbf{A}uto\textbf{R}egressive \textbf{G}eneration (\textit{\textbf{\modelname}}) framework, based on a novel \emph{holistic token sequence representation}. The core idea is to encode the complete geometry and topology of a B-rep as a single token sequence, enabling direct autoregressive modeling (see Figure~\ref{fig:BrepARG}). The token sequence representation comprises holistic tokenization and sequence construction (see Figure~\ref{fig:sequence}). For tokenization, geometry and topology of faces and edges are discretized separately, forming 3 types of tokens. The geometry primitives, i.e., faces and edges, are UV-sampled and tokenized into \textit{Geometry Tokens} by mapping their vector-quantized variational autoencoder (VQ-VAE) encoder latents to codebook indices via nearest neighbor search. The 3D positions, i.e., bounding boxes, are encoded into \textit{Position Tokens} through a newly designed \textit{uniform scalar quantization} algorithm. The topology information is explicitly represented through shared \textit{Face Index Tokens}. For sequence construction, we first construct geometry blocks, each consisting of all 3 types of tokens and representing a face or an edge. Then we use a topology-aware sequentialization scheme to sequence all face and edge blocks respectively, forming the face and edge block sequences, enforcing causal ordering while preserving local structural relationships. Finally, we assemble face block sequence and edge block sequence, together with necessary markers, to form the final holistic sequence representation.

To better leverage the powerful autoregressive framework, \textit{\modelname} then employs a transformer-based autoregressive model to learn the holistic geometric-topological token sequence via the next-token prediction task, thereby learning their joint distribution. This design enables the model to co-generate geometric shapes and topological connections in a single stream, achieving end-to-end autoregressive generation of B-rep sequences. Experiments demonstrate that \textit{\modelname} achieves state-of-the-art (SOTA) performance in B-rep generation while maintaining high efficiency, requiring only about 1.2 days to train on DeepCAD~\cite{deepcad} with 4 NVIDIA H20 GPUs and around 1.5 seconds per B-rep for inference on a single RTX 4090. With \modelname, we make the following contributions:
\begin{itemize}
    \item We propose a novel \textit{\textbf{holistic token sequence representation}} for B-rep, encoding both geometric and topological information into a single token sequence, addressing the long-standing fragmentation between geometry and topology in traditional approaches.
    \item We develop an \textit{\textbf{autoregressive framework}} for B-rep generation that holistically learns geometric and topological structures through self-attention, eliminating multi-stage pipelines and improving efficiency.
    \item We conduct experiments to show that \textit{\modelname} achieves SOTA performance on B-rep generation tasks.
\end{itemize}

\section{Related work}
\subsection{CAD Generation}
Early CAD generation methods predominantly used Constructive Solid Geometry (CSG)~\cite{Inversecsg,D2CSG,Csgnet,Csg-stump,UCSG-NET,Capri-net}, constructing models via Boolean operations on simple primitives, which limits their ability to represent complex geometries.

Another line of work~\cite{deepcad,skexgen,Secad-net,hnc-cad,MamTiff-CAD} learns parametric CAD command sequences that capture sketch structures, extrusion parameters, and action orderings to synthesize 3D shapes. Yet the development of this approach is currently limited by two main factors: 1) difficulty in supporting complex CAD design commands, 2) a persistent scarcity of high-quality, large-scale datasets.

\subsection{B-rep Generation}
B-rep represents objects by combining parametric geometric primitives with topological relationships, serving as a fundamental representation in industrial CAD. Prior work \cite{UV-net,NeuroNURBS,AAGNet,BRepGAT,FuS-GCN,brepformer} has studied B-rep classification and segmentation. For direct B-rep generation,~\cite{brepgen,solidgen} employ multistage models for geometry learning and use multi-level pointer networks or tree structures to incrementally construct topology. DTGBrepgen~\cite{dtgbrepgen} disentangles topology and geometry learning by employing separate networks to model topological structures and geometric primitives, respectively. \cite{brepgiff} encodes B-rep structure as a 3D graph for diffusion while still employing multipipeline encoders for geometric elements. 
BrepDiff~\cite{brepdiff} and Hola~\cite{hola} exclusively model face geometries, requiring additional steps to reconstruct the full B-rep topology.


\subsection{Autoregressive Models for 3D Generation}
In the field of large language models, the autoregressive next-token prediction paradigm has become the mainstream approach~\cite{gpt2,llama3,Scaling,Palm,Bloom,Opt}, demonstrating remarkable performance and scalability. Recently, this paradigm has also been extended to 3D content generation, where key advances rely on effective serialization of 3D data. Existing methods primarily focus on point clouds \cite{Pointgrow,Shapeformer,Pointgpt,Point-bert,Autosdf} and meshes~\cite{Polygen,meshgpt,G3pt,armesh,deepmesh}, using various ordering or encoding strategies for sequence modeling. 


For B-rep generation, early methods such as SolidGen~\cite{solidgen} model basic primitives in separate stages. Concurrent with our work, AutoBrep~\cite{autobrep} and BrepGPT~\cite{brepgpt} independently explore B-rep generation through sequence modeling. Both our method and AutoBrep employ dedicated tokens to encode topology, but they differ in sequence construction strategies and the implementation of discrete geometry representations. In addition to random rotation, we also apply re-indexing to improve the model’s generalization performance.

\section{Holistic Token Sequence Representation}
\label{sec:Representation}
The core challenge of holistic B-rep representation lies in the tight coupling between continuous, heterogeneous parametric geometry and discrete topology. To address this, we design a holistic token-sequence representation from two perspectives: holistic tokenization and sequence construction, as illustrated in Figure~\ref{fig:sequence}.

\subsection{Holistic Tokenization}
\label{sec:holistic}
To form a holistic tokenization for B-rep, we need to tokenize geometry primitives (i.e., faces and edges) and topology. Thus, the holistic tokenization of B-rep comprises three components: 
\textbf{\textit{1) Geometry Tokenization}}, which tokenizes faces and edges;
\textbf{\textit{2) Position Tokenization}}, which tokenizes the 3D positions (i.e., bounding boxes) of faces and edges in the B-rep for generation;
\textbf{\textit{3) Topology Tokenization}}, which encodes the topology of B-rep into tokens.

\paragraph{\textit{Geometry Tokenization}.} 
For geometric faces and edges, we construct learnable representations via regular sampling in the UV parameter domain. For each face, we build an evenly spaced UV grid and sample $32\times 32$ points, forming a discrete representation $\mathbf{F}\in\mathbb{R}^{32\times 32\times 3}$. For each edge, we uniformly sample $32$ points along the $U$-axis, forming a discrete representation $\mathbf{E}\in\mathbb{R}^{32\times 3}$. To reduce the geometry token vocabulary and enable token reuse, we employ a unified \textit{VQ-VAE} for both faces and edges based on a 2D convolutional U\text{-}Net architecture. To match the input dimensionality required by the 2D convolutional network, $\mathbf{E}$ is broadcast to $\mathbf{E}^{\prime}\in\mathbb{R}^{32\times32\times3}$. To balance quantization fidelity and codebook size, we downsample the sampled features by a factor of 16, yielding a $2\times2$ latent feature map composed of four latent vectors. Each latent vector is then mapped to the nearest codeword in the codebook through a nearest neighbor search, and the corresponding indices are used as quantized representations. Consequently, the four indices derived from each face and edge constitute our final \textit{Geometry Tokens}.

\paragraph{\textit{Position Tokenization}.}
It is crucial to provide 3D positional information for geometric primitives (i.e., faces and edges) during B-rep generation. However, the large number of bounding boxes makes it difficult to tokenize the positional data, which are represented by six scalar values $\mathbf{b}=[x_{\min},y_{\min},z_{\min},x_{\max},y_{\max},z_{\max}]\in[-1,1]^6$. We find that when a single codeword is used to represent an entire bounding box, the \textit{VQ-VAE} degenerates into an $n$-means clustering process over a large sample space, thus failing to achieve sufficient geometric accuracy. We also observe that splitting a bounding box into multiple subvectors and quantizing them into separate codewords still yields limited reconstruction fidelity, as the large data volume makes it difficult for the model to learn a precise mapping from continuous coordinates to discrete tokens.

To achieve a more stable mapping to token representations and higher quantization precision, we tokenize bounding boxes using coordinate-wise \textit{uniform scalar quantization}, where each coordinate is quantized using mathematical formulas to directly map the values to discrete token indices. Specifically, for each coordinate $b_j$ ($j\!\in\!\{1,\dots,6\}$), we first normalize and clip it to $[0,1]$ to obtain $\tilde{b}_j$. We then scale it onto an $L$-level uniform grid and round to the nearest integer to obtain a discrete index:
\begin{equation}
k_j = \mathrm{round}\!\big((L-1)\,\tilde{b}_j\big)\;\in\;\{0,\ldots,L-1\}.
\label{equ:bbox tokenize}
\end{equation}
This procedure is applied independently to all six coordinates, yielding 6 \textit{Position Tokens} $(k_1,\ldots,k_6)$. The corresponding dequantization uses a linear mapping:
\begin{equation}
b_j = \frac{2\,k_j}{L-1} - 1.
\label{equ:bbox detokenize}
\end{equation}

\paragraph{\textit{Topology Tokenization}.} 
The essence of topological relationships lies in representing the connectivity among geometry primitives. Therefore, as long as connectivity can be expressed, topology can be tokenized. Thus, our topology tokenization strategy employs a labeling approach on tokens derived from geometric primitives to encode the adjacency relationships between edges and faces.

Specifically, all closed face regions and boundary edges are split along their seams to ensure that each edge serves as the boundary between two distinct faces, and that the endpoints of each edge correspond to different vertices. By implicitly representing vertices through edge endpoints, we focus on faces and their adjacency topology. Firstly, we append a face index to each face as a unique identifier for subsequent references. Then, we append two face indices to each edge to explicitly encode face-edge adjacency.

\subsection{Sequence Construction}
We construct sequence representation hierarchically. We first construct geometry blocks for each face and edge with previously tokens. Then, two types of geometry blocks are respectively ordered with designed algorithms. Finally, we assemble the final holistic sequence representation.

\paragraph{\textit{Geometry Block Construction}.}
\label{sec:block}
We design two geometry blocks respectively for faces and edges. Each \textit{face block} $f_i$ contains three types of tokens: 6 consecutive Position Tokens $t^p_i$, 4 consecutive Geometry Tokens $t^g_i$, and a \textit{Face Index Token} $t^{idx}$, formulated as:

\begin{equation}
f_i = [(t^p_1,...,t^p_6), \;(t^g_1,...,t^g_4), \;t^{idx}]
\end{equation}

Each \textit{edge block} $e_j$ begins with 2 \textit{Face Index Tokens} $t^{idx}_k$, corresponding to the pair of faces that share this edge. In this way, we explicitly encode the topology, i.e., the connectivity of two faces and one edge. The \textit{Face Index Tokens} are followed by the 6 Position tokens $t^p_i$ and 4 Geometry Tokens $t^g_j$. The whole \textit{edge block} is formulated as:

\begin{equation}
e_j = [t^{idx}_1, \;t^{idx}_2, \;(t^p_1,...,t^p_6), \;(t^g_1,...,t^g_4)]
\end{equation}


\begin{figure}[t]
    \centering
    \includegraphics[width=\columnwidth]{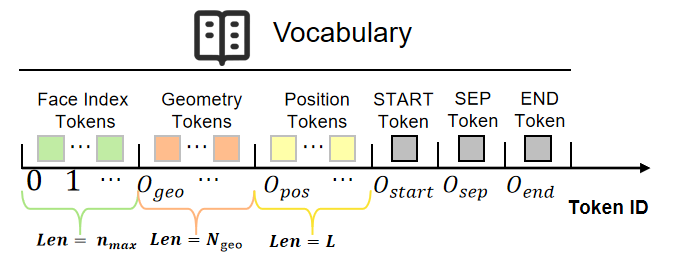}
    \caption{Unified vocabulary. Face Index, Geometry, Position, and Special tokens (\textit{START}, \textit{SEP}, \textit{END}) are unified into a nonoverlapping vocabulary via predefined offsets.}
    \label{fig:vocabulary}
\end{figure}

\begin{figure*}[t]
    \centering
    \includegraphics[width=\textwidth]{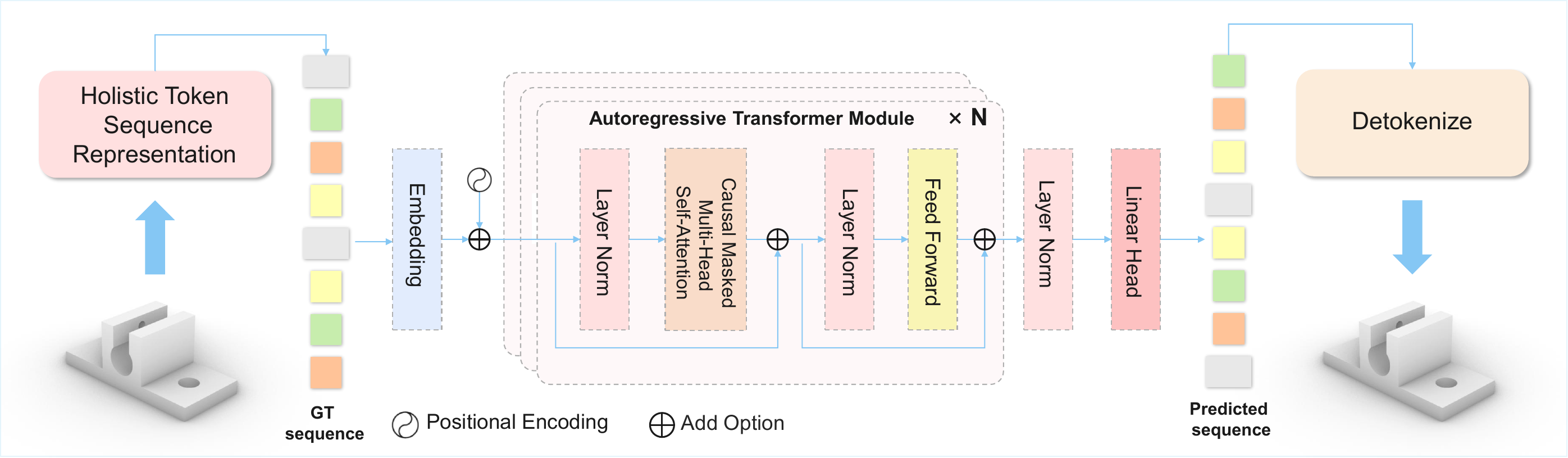}
    \caption{B-rep generation pipeline using a sequential generative model. The holistic token sequence of the B-rep is fed into an autoregressive, decoder-only Transformer for generation, and the predicted token sequence is detokenized to reconstruct the complete B-rep. 
    }
    \label{fig:Generative Model}
\end{figure*}

\paragraph{\textit{Geometry Block Sequentialization}.}
In this section, we design a topology-aware sequentialization strategy for sequencing \textit{face blocks} and \textit{edge blocks}, respectively. When sequentializing geometric primitives with embedded topological information, arbitrary ordering may compromise the structural integrity of the original B-rep model. To mitigate this issue, we construct a causal and topology-aware sequence order. Faces are ordered by first selecting the highest-degree face and then performing a depth-first search (DFS) that prioritizes lower-degree neighbors, so as to place topologically adjacent faces closer together in the sequence, yielding the \textit{face block} sequence \(S_f\). 
Edges are then ordered according to the maximum adjacent face index in ascending order (MAX-IDX-A), ensuring that edges are placed near their associated faces, which tightens the attention span between faces and edges and alleviates long-range dependencies, forming the \textit{edge block} sequence \(S_e\). After sequencing, we follow GraphGPT~\cite{graphgpt} and re-index all \textit{Face Index Tokens}, where a random integer \(r \in [0, n_{\text{max}})\) is added and then taken modulo \(n_{\text{max}}\), with \(n_{\text{max}}\) denoting the maximum face count across all B-rep models in the dataset. This operation randomizes the topology encoding for generalization.

\paragraph{\textit{Final Sequence Assembly}.} 
To simplify the decoding process during detokenization, we organize the holistic token sequence so that all face blocks are enumerated first, followed by all edge blocks, which can be defined as:

\begin{equation}
\mathcal{S} = [\, \text{\textit{START}},\; S_f,\; \text{\textit{SEP}},\; S_e,\; \text{\textit{END}} \,]
\label{eq:sequence_structure}
\end{equation}

Here, \textit{START}, \textit{SEP}, and \textit{END} are \textit{Special Tokens} denoting the sequence start, the separator between face and edge subsequences, and the sequence end, respectively. $S_f$ and $S_e$ represent all face and edge blocks, respectively.

To embed discrete symbols from different sources, including \textit{Face Index Tokens}, \textit{Geometry Tokens}, \textit{Position Tokens}, and \textit{Special Tokens}, into a unified space without overlap, we adopt a segmented, non-overlapping linear token allocation scheme. In this design, each token subset is assigned a distinct starting offset along the global token index range, ensuring that all token types occupy disjoint index intervals. Specifically, let $n_{\max}$ denote the maximum number of B-rep faces in the dataset, $N_{\mathrm{geo}}$ the size of the \textit{VQ-VAE} codebook, and $L$ the number of scalar quantization levels for bounding boxes. The segment offsets are defined as $o_{\mathrm{geo}}=n_{\max}$, $o_{\mathrm{pos}}=n_{\max}+N_{\mathrm{geo}}$, and $o_{\mathrm{spec}}=n_{\max}+N_{\mathrm{geo}}+L$. This scheme merges the sub-vocabularies into a single continuous token space without index conflicts, as illustrated in Figure~\ref{fig:vocabulary}.

\section{Generation Framework}

The model \textit{\modelname} consists of two main components:
1) a \textit{VQ-VAE} that compresses and quantizes the UV-sampled features of faces and edges into a shared codebook, and 2) an autoregressive training process based on a decoder-only transformer under teacher forcing, using the holistic token sequence representation (see Section~\ref{sec:Representation} for details). Finally, we introduce a post-processing algorithm that detokenizes the generated sequences to reconstruct the complete B-rep model.

\subsection{Face-Edge VQ-VAE}
To obtain discrete representations of geometric information for faces and edges, we employ a \textit{VQ-VAE} model built upon a 2D convolutional U\text{-}Net backbone. Given an input $\mathbf{x}\in\mathbb{R}^{32\times32\times3}$ (geometric features of faces or edges, $\mathbf{F}$ or $\mathbf{E}^{\prime}$; see  Section~\ref{sec:holistic}), the encoder downsamples it by a factor of 16 to produce a latent feature map $\mathbf{z}_e\in\mathbb{R}^{2\times2\times128}$. Then a $1{\times}1$ pointwise convolution linearly projects $\mathbf{z}_e$ to 64 channels, yielding $\mathbf{z}'_e\in\mathbb{R}^{2\times2\times64}$ to match the dimensionality of the codebook. Each latent vector in $\mathbf{z}'_e$ is then mapped to its nearest codeword in the codebook through a nearest neighbor search, producing the quantized representation $\mathbf{z}_q\in\mathbb{R}^{2\times2\times64}$. The decoder reconstructs the input features as $\hat{\mathbf{x}} = D(\mathbf{z}_q)$, and the reconstruction loss is computed between $\mathbf{x}$ and $\hat{\mathbf{x}}$ as $\mathcal{L}_{\mathrm{rec}} = \lVert \mathbf{x} - \hat{\mathbf{x}} \rVert_2^2$.

To mitigate codebook collapse during training, we follow the codebook restart strategy of \textit{CVQ-VAE}~\cite{CVQ}: a feature-pool-driven online reinitialization together with probabilistic nearest neighbor sampling is used to revive low-usage codebook entries. This design improves codebook utilization and provides a more robust discrete representation for subsequent autoregressive modeling.

\subsection{Sequential Generative Model}
To effectively capture the geometric and topological regularities within holistic token sequences, we employ a decoder-only transformer architecture equipped with causal masking to perform autoregressive next-token prediction, as illustrated in Figure~\ref{fig:Generative Model}. Specifically, we define the training dataset as $\mathcal{D}=\{\mathbf{S}^{(n)}\}_{n=1}^{N}$, where $N$ is the total number of training sequences and each $\mathbf{S}=(t_1, t_2, \ldots, t_T)$ contains $T$ discrete tokens $t_i \in V$, with $V$ denoting the token vocabulary. Each token is embedded into a $d$-dimensional vector with added discrete positional information, yielding a representation $\mathbf{e}(t_i)\in\mathbb{R}^d$. The resulting embeddings are processed by a decoder-only transformer backbone composed of stacked multi-head self-attention and feed-forward layers, which predicts the next token from the vocabulary conditioned on the preceding context. The transformer model, parameterized by $\theta$, defines conditional probabilities $p_{\theta}(t_i \mid \mathbf{e}(t_{<i}))$, and the overall training objective is to maximize the joint probability of all sequences in the dataset:
\begin{equation}
\prod_{\mathbf{S}\in\mathcal{D}}\;\prod_{i=1}^{T(\mathbf{S})}
p_{\theta}\!\left(t_i \mid \mathbf{e}(t_{<i});\,\theta\right),
\label{eq:autoregressive_training}
\end{equation}

At inference time, tokens are generated autoregressively by sampling from $p_{\theta}(t_i \mid \mathbf{e}(t_{<i}))$, starting from the~\textit{START} token and continuing until the \textit{END} token is produced.

\subsection{Holistic Token Sequence Detokenization}
For each generated holistic token sequence, the regular token-length design and SEP delimiters enable direct extraction of token blocks corresponding to individual faces and edges. Subsequently, we apply the VQ-VAE decoder and Equation \ref{equ:bbox detokenize} to detokenize the \textit{Geometry Tokens} and \textit{Position Tokens}, respectively, thereby recovering their geometric representations. Face-edge adjacency relationships are established via shared \textit{Face Index Tokens}, which further enables the construction of face–face adjacency (as each edge connects two faces).
To infer vertex information and reconstruct the B-rep model, we introduce a novel post-processing algorithm. First, we treat the endpoints of each edge as candidate vertices. Then, using a union-find-based greedy clustering approach, we iteratively merge the closest candidate points from different edges within each face boundary based on geometric proximity and local loop topology, while ensuring global consistency by unifying shared vertex groups across faces. Finally, each vertex is determined as the centroid of its constituent candidate points, resulting in a complete and unique vertex list, edge-vertex connectivity, and loop structures. The resulting geometry and topology are seamlessly stitched into a valid B-rep solid using OpenCascade's sew function.


\begin{figure*}[t]
    \centering
    \begin{subfigure}[t]{0.32\textwidth}
        \centering
        \includegraphics[width=\linewidth]{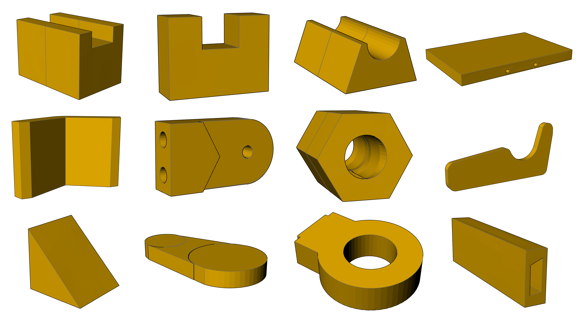}
        \caption*{BrepGen}
    \end{subfigure}
    \hfill
    \begin{subfigure}[t]{0.32\textwidth}
        \centering
        \includegraphics[width=\linewidth]{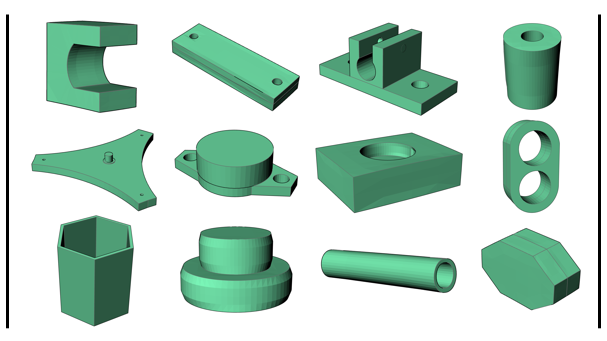}
        \caption*{DTGBrepGen}
    \end{subfigure}
    \hfill
    \begin{subfigure}[t]{0.32\textwidth}
        \centering
        \includegraphics[width=\linewidth]{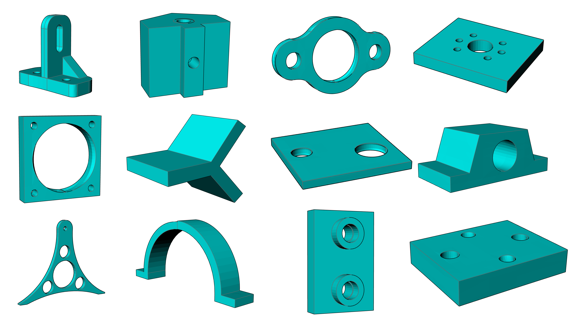}
        \caption*{Ours}
    \end{subfigure}
    
    \caption{Qualitative comparison of B-rep models generated by  BrepGen~\cite{brepgen}, DTGBrepgen~\cite{dtgbrepgen}, and our method (\textit{\modelname}) on the DeepCAD dataset.}
    \label{fig:brep_comparison}
\end{figure*}

\section{Experiment}

\subsection{Experiment Setup}
\textbf{Datasets.}
We use three datasets for training and evaluation. The DeepCAD dataset~\cite{deepcad} is employed for unconditional generation, ablation studies, and efficiency evaluation, while the Furniture dataset~\cite{brepgen} is used for class-conditioned generation. Both datasets are filtered following the same strategy as DTGBrepGen~\cite{dtgbrepgen}, removing B-rep models with more than 50 faces or with any single face containing over 30 edges. After filtering, the resulting sets contain 80,509 DeepCAD B-reps and 1,065 Furniture B-reps. ABC dataset~\cite{Abc} is used for unconditional generation. We extend the above filtering criteria by additionally excluding simple CAD models with fewer than 10 faces, yielding 105,798 valid B-reps.

\vspace{0.5em}
\noindent\textbf{Network architecture.} 
\emph{Quantization \& codebooks.}
We apply \textit{uniform scalar quantization} (see Equation~\ref{equ:bbox tokenize}) with $L=$2,048 levels. The \textit{VQ-VAE} uses a codebook of size 4,096 on the DeepCAD dataset, 8,192 for the more complex ABC dataset. The network comprises 5 convolutional encoder blocks and 5 symmetric decoder blocks: the first four encoder blocks perform downsampling sequentially, while the last block extracts features. The output channels at each stage are 32, 64, 128, 256, and 512, enabling progressive compression and abstraction of the input features.

\emph{Sequential generative model.}  
For sequence generation, we use a decoder-only transformer with 8 layers and 8 attention heads. Each layer has an embedding dimension of 256 and a feed-forward hidden dimension of 1,024.

\vspace{0.5em}
\noindent\textbf{Training.}  
\emph{VQ-VAE.}
During the VQ-VAE training phase, we use the AdamW optimizer~\cite{loshchilov2017decoupled}  with a learning rate of \(1\times10^{-4}\), a weight decay of \(1\times10^{-6}\), and batch sizes of 2,048 and 8,192 for the DeepCAD and ABC datasets, respectively. To improve training efficiency and stability, we utilize automatic mixed precision (AMP) and DistributedDataParallel (DDP) for multi-GPU parallelization. On the DeepCAD dataset, the model is trained on 4 NVIDIA H20 GPUs, requiring approximately 12 hours.


\emph{Sequential generative model.}  
During the sequential generative model training phase, we also use the AdamW optimizer with a learning rate of $1\times10^{-3}$, weight decay of $1\times10^{-2}$, and a batch size of 128. AMP and DDP are again employed for efficient multi-GPU training. On the DeepCAD dataset, the model is trained for 500 epochs using 4 NVIDIA H20 GPUs, which takes approximately 17 hours.

\vspace{0.5em}
\noindent\textbf{Inference.}  
We adopt a probabilistic autoregressive sampling strategy to generate B-rep sequences. The generation process begins with an initial context prompt: unconditional generation starts from the \textit{START} token, whereas conditional generation uses a category-specific token. At each iteration, the model predicts a probability distribution over the vocabulary conditioned on all previously generated tokens. We apply nucleus (top-$p$) sampling~\cite{holtzman2019curious}, which selects the next token from the smallest subset of the vocabulary whose cumulative probability is at least $p$. The sampling process terminates when the \textit{END} token is generated or when the maximum sequence length is reached.

\vspace{0.5em}
\noindent\textbf{Evaluation metrics.}  
Following BrepGen~\cite{brepgen}, we adopt both distributional and CAD-specific metrics for comprehensive evaluation. Distribution metrics include Coverage (COV), Minimum Matching Distance (MMD), and Jensen-Shannon Divergence (JSD). COV measures the proportion of reference samples that are matched by generated samples, based on nearest-neighbor matching with Chamfer distance. MMD computes the average Chamfer distance from each reference sample to its closest generated sample. JSD quantifies the Jensen-Shannon divergence between point cloud histograms of generated and reference sets.  
CAD metrics include Novelty, Uniqueness, and Validity. Novelty measures the proportion of generated samples not present in the training set hash table. Uniqueness evaluates the ratio of distinct samples after deduplication within the generated set. Validity measures the proportion of watertight, topologically consistent, and non-degenerate samples verified by a CAD kernel.


\begin{table}[t]
\centering
\caption{Comparison of unconditional generation results on the DeepCAD and ABC datasets. (MMD and JSD values are multiplied by $10^{2}$; COV, Novel, Unique, and Valid are percentages.)}
\setlength{\tabcolsep}{1.0pt}
\small
\begin{tabular}{lcccccc}
\toprule
\textbf{Method} & \multicolumn{6}{c}{\textbf{DeepCAD}} \\
\cmidrule(lr){2-7}
& \textbf{COV~$\uparrow$} & \textbf{MMD~$\downarrow$} & \textbf{JSD~$\downarrow$} & \textbf{Novel~$\uparrow$} & \textbf{Unique~$\uparrow$} & \textbf{Valid~$\uparrow$} \\
\midrule
DeepCAD     & 70.81 & 1.31 & 1.79 & 93.80 & 89.79 & 58.10 \\
BrepGen     & 72.38 & 1.13 & 1.29 & 99.72 & 99.18 & 68.23 \\
DTGBrepGen  & 74.52 & 1.07 & 1.02 & 99.79 & 98.94 & 79.80 \\
BrepDiff  &69.73  &0.93 &1.34  &99.76  &99.40  & 62.83 \\
\textbf{Ours} & \textbf{75.45} & \textbf{0.89} & \textbf{1.02} & \textbf{99.82} & \textbf{99.80} & \textbf{87.60} \\
\midrule
\textbf{Method} & \multicolumn{6}{c}{\textbf{ABC}} \\
\cmidrule(lr){2-7}
& \textbf{COV~$\uparrow$} & \textbf{MMD~$\downarrow$} & \textbf{JSD~$\downarrow$} & \textbf{Novel~$\uparrow$} & \textbf{Unique~$\uparrow$} & \textbf{Valid~$\uparrow$} \\
\midrule
DTGBrepGen  & 66.07 & 1.456 & 1.757 & 99.43 & 99.27 & 57.59 \\
\textbf{Ours} & \textbf{70.10} & \textbf{1.405} & \textbf{1.337} & \textbf{99.78} & \textbf{99.72} & \textbf{67.54} \\
\bottomrule
\end{tabular}
\label{tab:deepcad_abc}
\vspace{-3pt}
\end{table}

\begin{table}[b]
\centering
\caption{Comparison of training and inference efficiency.}
\setlength{\tabcolsep}{3.5pt}
\small
\begin{tabular}{lcccc}
\toprule
\textbf{Method} &
\makecell{\textbf{Training}\\\textbf{Time}} &
\makecell{\textbf{Training}\\\textbf{GPU}} &
\makecell{\textbf{Inference}\\\textbf{Time}} &
\makecell{\textbf{Inference}\\\textbf{GPU}} \\
\midrule
BrepGen     & 7.5 days    & H20 $\times$ 4 & 8.4 s  & RTX4090 $\times$ 1 \\
DTGbrepgen  & 3.0 days    & H20 $\times$ 4 & 3.6 s    & RTX4090 $\times$ 1 \\
BrepDiff  & 1.8 days    & H20 $\times$ 4 & 8.3 s    & RTX4090 $\times$ 1 \\
\textbf{Ours} & \textbf{1.2 days} & H20 $\times$ 4 & \textbf{1.5 s} & RTX4090 $\times$ 1 \\
\bottomrule
\end{tabular}
\label{tab:efficiency}
\vspace{-4pt}
\end{table}

\subsection{Unconditional Generation}
To evaluate the performance of our model on unconditional generation, we conduct a comprehensive quantitative comparison on the DeepCAD dataset. For distribution metrics, we compute statistics over 3,000 generated B-rep models and 1,000 reference models, where each model is represented as a point cloud obtained by uniformly sampling 2,000 points. CAD metrics are directly evaluated on the 3,000 generated B-rep entities. 

We report the mean of 10 independent runs, as shown in Table~\ref{tab:deepcad_abc}. Using nucleus (top-$p$) sampling with $p = 0.9$ for DeepCAD and $p = 0.8$ for ABC, our approach consistently outperforms all baseline methods across all evaluation metrics. Notably, our model achieves a validity score of $87.6\%$ on the DeepCAD dataset, demonstrating its strong ability to capture the geometric and topological structures of B-reps. As shown in Figure~\ref{fig:brep_comparison}, qualitative comparisons further reveal that \textit{\modelname} generates more realistic and geometrically accurate B-rep models than the baselines.

Beyond generation quality, our generative model also demonstrates significant advantages in training and inference efficiency, as detailed in Table~\ref{tab:efficiency}. Furthermore, we analyze the effect of different sampling strategies on generation outcomes. As shown in Table~\ref{tab:top_p_ablation}, by adjusting the $p$ parameter in nucleus sampling, we can flexibly balance between generating more diverse models and achieving a higher proportion of valid B-rep outputs.
\begin{table}[t]
\centering
\caption{Results under different nucleus sampling thresholds $p$. Note that MMD and JSD values are multiplied by $10^{2}$; COV, Novel, Unique, and Valid are percentages.}
\small
\setlength{\tabcolsep}{2.5pt} 
\resizebox{\linewidth}{!}{
\begin{tabular}{ccccccc}
\toprule
\textbf{$p$} & 
\textbf{COV $\uparrow$} & 
\textbf{MMD $\downarrow$} & 
\textbf{JSD $\downarrow$} & 
\textbf{Novel $\uparrow$} & 
\textbf{Unique $\uparrow$} & 
\textbf{Valid $\uparrow$} \\
\midrule
0.95  & \textbf{76.10} & \textbf{0.8779} & 1.023 & \textbf{99.92} & \textbf{99.88} & 82.85 \\
0.9  & 75.45 & 0.8869 & \textbf{1.017} & 99.82 & 99.80 & 87.60 \\
0.8  & 73.71 & 0.9142 & 1.360 & 99.79 & 99.61 & 89.23 \\
0.7  & 69.81 & 0.9594 & 2.370 & 99.21 & 98.62 & 89.51 \\
0.6  & 63.50 & 1.0180 & 4.260 & 98.92 & 95.13 & \textbf{90.25} \\
\bottomrule
\end{tabular}
}
\label{tab:top_p_ablation}
\vspace{-3pt}
\end{table}

\begin{figure*}[t]
    \centering
    \begin{subfigure}[t]{0.09\textwidth}
        \centering
        \includegraphics[width=\linewidth]{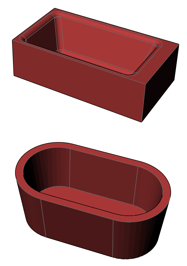}
        \caption*{batchtub}
    \end{subfigure}
    \hfill
    \begin{subfigure}[t]{0.09\textwidth}
        \centering
        \includegraphics[width=\linewidth]{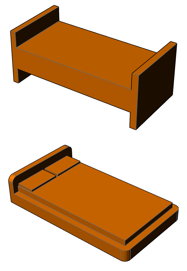}
        \caption*{bed}
    \end{subfigure}
    \hfill
    \begin{subfigure}[t]{0.09\textwidth}
        \centering
        \includegraphics[width=\linewidth]{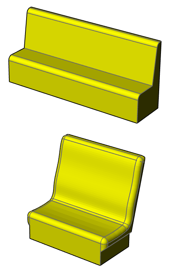}
        \caption*{bench}
    \end{subfigure}
    \hfill
    \begin{subfigure}[t]{0.09\textwidth}
        \centering
        \includegraphics[width=\linewidth]{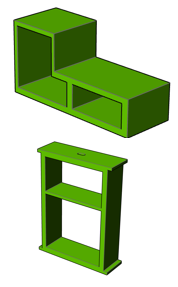}
        \caption*{bookshelf}
    \end{subfigure}
    \hfill
    \begin{subfigure}[t]{0.09\textwidth}
        \centering
        \includegraphics[width=\linewidth]{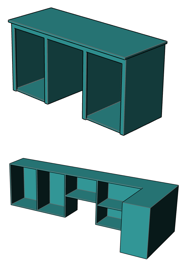}
        \caption*{cabinet}
    \end{subfigure}
    \hfill
    \begin{subfigure}[t]{0.09\textwidth}
        \centering
        \includegraphics[width=\linewidth]{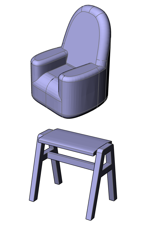}
        \caption*{chair}
    \end{subfigure}
    \hfill
    \begin{subfigure}[t]{0.09\textwidth}
        \centering
        \includegraphics[width=\linewidth]{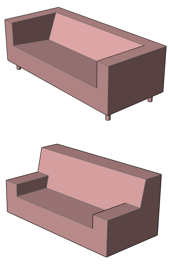}
        \caption*{couch}
    \end{subfigure}
    \hfill
    \begin{subfigure}[t]{0.09\textwidth}
        \centering
        \includegraphics[width=\linewidth]{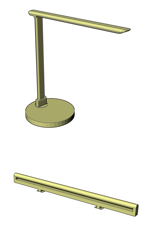}
        \caption*{lamp}
    \end{subfigure}
    \hfill
    \begin{subfigure}[t]{0.09\textwidth}
        \centering
        \includegraphics[width=\linewidth]{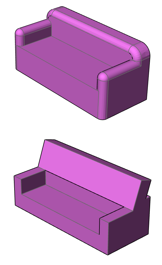}
        \caption*{sofa}
    \end{subfigure}
    \hfill
    \begin{subfigure}[t]{0.09\textwidth}
        \centering
        \includegraphics[width=\linewidth]{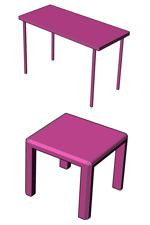}
        \caption*{table}
    \end{subfigure}
    \hfill
    
    \caption{Class-conditioned generation. We showcase two representative B-rep samples for each category in the Furniture dataset, demonstrating our model's ability for conditional generation.}
    \label{fig:Class-conditioned generation}
\end{figure*}

\begin{figure}[htbp]
    \centering
    \includegraphics[width=\columnwidth]{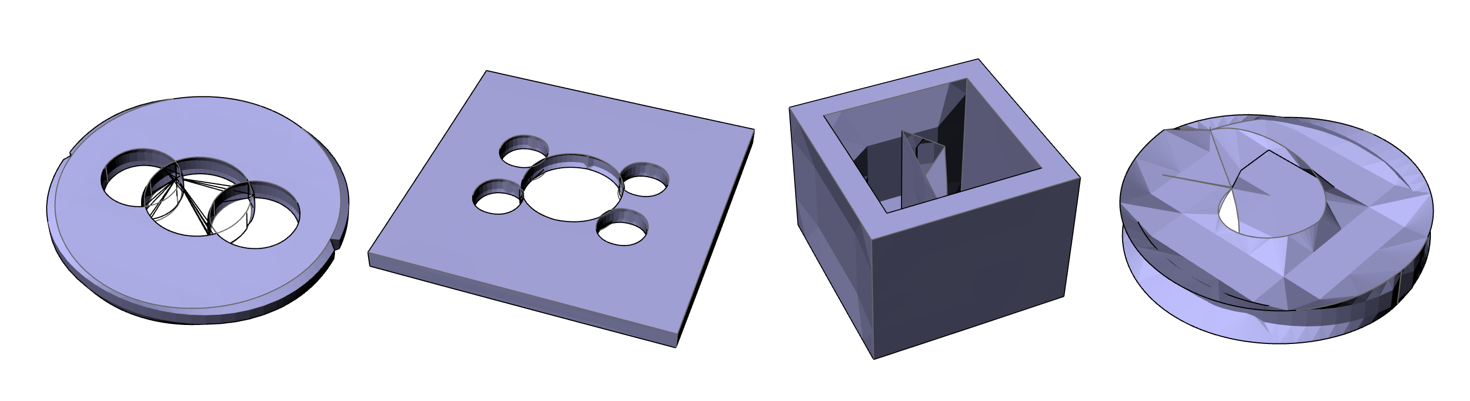}
    \caption{Failure cases generated by our method (\textit{\modelname})}
    \label{fig:Invalid}
\end{figure}

\subsection{Class-conditioned Generation}

We demonstrate the class-conditional generation capability of \textit{\modelname} using a prefix-based conditioning strategy. During training, the \textit{START} token at the beginning of each sequence is replaced with a class-specific token, enabling the model to learn class-aware autoregressive generation. At inference time, conditional generation is performed by prefixing the input sequence with the corresponding class token. Additional qualitative results are shown in Figure~\ref{fig:Class-conditioned generation}.

\subsection{Failure Generation Cases}

As shown in Figure~\ref{fig:Invalid}, we observe several failure cases, which can be attributed to two primary factors: 1) precision loss introduced by the \textit{VQ-VAE} during quantization of geometric features of faces and edges, and 2) the increased complexity of modeling long autoregressive sequences.
Future work will explore higher-fidelity geometric quantization schemes and more efficient autoregressive modeling strategies to further improve the stability and consistency of generated results.

\begin{table}[b]
\centering
\caption{Ablation study of face ordering strategies. Note that MMD and JSD values are multiplied by $10^{2}$; COV, Novel, Unique, and Valid are percentages.}
\setlength{\tabcolsep}{3pt} 
\small
\setlength{\tabcolsep}{2.0pt}
\begin{tabular}{lcccccc}
\toprule
\textbf{Method} & \textbf{COV~$\uparrow$} & \textbf{MMD~$\downarrow$} & \textbf{JSD~$\downarrow$} & \textbf{Novel~$\uparrow$} & \textbf{Unique~$\uparrow$} & \textbf{Valid~$\uparrow$} \\
\midrule
RAND  & 71.10 & 0.9473 & 1.128 & 99.27 & 99.67 & 67.92 \\
ZYX   & 74.82 & 0.9165 & 1.018 & 99.73 & 99.69 & 83.12 \\
DEG-A & 74.09 & 0.9059 & 1.290 & 99.52 & 99.35 & 82.98 \\
SS    & 72.91 & 0.9551 & 1.403 & 99.58 & 99.19 & 79.38 \\
BFS   & 74.65 & 0.9175 & 1.128 & 99.67 & 99.73 & 85.94 \\
\textbf{Ours} & \textbf{75.45} & \textbf{0.8869} & \textbf{1.017} & \textbf{99.82} & \textbf{99.80} & \textbf{87.60} \\
\bottomrule
\end{tabular}
\label{tab:face_order_ablation}
\end{table}

\subsection{Ablation Studies}
\label{Ablation Studies}
To verify the effectiveness of the topology-aware sequentialization strategy in sequence construction, we conducted a systematic ablation study to evaluate various heuristic ordering schemes, including face ordering and edge ordering.

For face ordering, we evaluate 6 strategies covering different design principles, while keeping edge ordering fixed to the maximum adjacent face index ascending (MAX-IDX-A) scheme for controlled comparison. Among topology-aware methods, we evaluate degree-ascending ordering (DEG-A), centroid-based ordering (ZYX; faces are sorted lexicographically by centroid coordinates in Z, then Y, then X), and spectral ordering (SS). We also explore traversal strategies starting from the highest-degree face, comparing breadth-first search (BFS) and our adopted depth-first search (DFS), both of which prioritize visiting low-degree neighbors. Results under consistent training and inference settings ($p = 0.9$) are summarized in Table~\ref{tab:face_order_ablation}.

For edge ordering, we compare random ordering (RAND) with the proposed MAX-IDX-A strategy, in which each edge is associated with the indices of its two adjacent faces, $i$ and $j$, and sorted in ascending order of $\max(i, j)$, with ties broken randomly. Results under consistent face ordering, training, and inference settings ($p = 0.9$) are summarized in Table~\ref{tab:edge_order_ablation}.

The results show that the DFS-based traversal in face ordering and the MAX-IDX-A strategy in edge ordering jointly contribute to the overall performance. These findings highlight that the proposed topology-aware sequentialization effectively captures local connectivity patterns, leading to more coherent geometry and stable generation.

\begin{table}[t]
\centering
\caption{Ablation study of edge ordering strategies. 
Note that MMD and JSD values are multiplied by $10^{2}$; COV, Novel, Unique, and Valid are percentages.}
\small
\setlength{\tabcolsep}{3pt} 
\resizebox{\linewidth}{!}{
\begin{tabular}{lcccccc}
\toprule
\textbf{Method} & 
\textbf{COV~$\uparrow$} & 
\textbf{MMD~$\downarrow$} & 
\textbf{JSD~$\downarrow$} & 
\textbf{Novel~$\uparrow$} & 
\textbf{Unique~$\uparrow$} & 
\textbf{Valid~$\uparrow$} \\
\midrule
RAND  & 74.25 & 0.913  & 1.073 & 99.49 & 99.71 & 85.43 \\
\textbf{Ours} & \textbf{75.45} & \textbf{0.887} & \textbf{1.017} & \textbf{99.82} & \textbf{99.80} & \textbf{87.60} \\
\bottomrule
\end{tabular}
}
\label{tab:edge_order_ablation}
\vspace{-3pt}
\end{table}

\section{Conclusion}
We propose \textit{\modelname}, the first framework that encodes the heterogeneous geometry and topology of B-rep into a holistic token sequence, reformulating B-rep generation as a sequence modeling task. This design enables sequence-based generation architectures, such as transformer, to jointly learn geometric details and topological constraints in a single process, effectively reducing multi-stage errors and computational overhead. Experimental results demonstrate that \textit{\modelname} achieves SOTA performance, paving a new direction for generative B-rep modeling.

\section*{Acknowledgements}
This work was supported by the National NaturalScience Foundation of China under Grant 62506302.

{
    \small
    \bibliographystyle{ieeenat_fullname}
    \bibliography{main}
}


\input{sec/X_suppl}

\end{document}

%% file: sec/X_suppl.tex
\clearpage
\setcounter{page}{1}
\appendix

\section{Appendix}
\subsection{Background of VQ-VAE}
We use the VQ-VAE to obtain discrete representations for faces and edges in the B-rep. Given a geometric input $\mathbf{x} \in \mathbb{R}^{H \times W \times C}$, the VQ-VAE learns to encode it using a set of discrete codebook vectors drawn from a learnable codebook. Specifically, the encoder $E_{\phi}$ maps $\mathbf{x}$ to a feature map $\mathbf{z}_{e} \in \mathbb{R}^{h \times w \times n_q}$, where $h$ and $w$ denote the spatial resolution of the downsampled feature map, and $n_q$ is the dimensionality of each latent vector. Each element $\mathbf{z}_{e,ij}$ in this feature map represents the latent vector located at spatial position $(i,j)$. To obtain a discrete representation, every latent vector $\mathbf{z}_{e,ij}$ is quantized to its nearest entry in a codebook 
$Z = \{\,\mathbf{e}_k\,\}_{k=1}^{K}$, 
where $K$ is the size of the codebook and each codeword $\mathbf{e}_k \in \mathbb{R}^{n_q}$ is a learnable prototype vector in the latent space. The quantization function $q(\cdot)$ replaces each latent vector with its closest codeword according to the squared Euclidean distance:
\begin{equation}
\mathbf{z}_{q,ij} = q(\mathbf{z}_{e,ij}) = \mathbf{e}_k, 
\quad 
k = \arg\min_{\mathbf{e}_k \in Z} \Vert \mathbf{z}_{e,ij} - \mathbf{e}_k \Vert_2^2
\label{eq:vq}
\end{equation}
This process effectively embeds the encoder output feature tensor into a discrete index grid of size $h \times w$, where each grid element corresponds to one codebook entry. The resulting discrete latent grid is then used by the decoder for reconstruction. The overall forward pass of the VQ-VAE is given by:
\begin{equation}
\hat{\mathbf{x}} = D_{\theta}(\mathbf{z}_q) 
= D_{\theta}(q(\mathbf{z}_{e})) 
= D_{\theta}(q(E_{\phi}(\mathbf{x})))
\label{eq:vqvae_forward}
\end{equation}
where $E_{\phi}$ and $D_{\theta}$ denote the encoder and decoder, respectively. The encoder maps the input to a continuous latent feature map, which is then quantized to the nearest entries in a learnable codebook, yielding discrete latent codes. The decoder reconstructs the input from these quantized representations.

\vspace{0.3em}
\noindent\textbf{Training Objective.}
During training, the encoder $E_{\phi}$, decoder $D_{\theta}$, and codebook $Z$ are jointly optimized by minimizing the following loss:
\begin{equation}
\mathcal{L} =
\Vert \mathbf{x} - \hat{\mathbf{x}} \Vert_2^2
+ \Vert \mathrm{sg}[E_{\phi}(\mathbf{x})] - \mathbf{z}_q \Vert_2^2
+ \beta \, \Vert E_{\phi}(\mathbf{x}) - \mathrm{sg}[\mathbf{z}_q] \Vert_2^2
\label{eq:vqvae_loss}
\end{equation}
where $\mathrm{sg}[\cdot]$ denotes the \textit{stop-gradient} operator, which prevents gradients from backpropagating through its argument.
The loss comprises three components:
\begin{itemize}
    \item \textbf{Reconstruction loss} $\Vert \mathbf{x} - \hat{\mathbf{x}} \Vert_2^2$, ensuring that the reconstructed sample $\hat{\mathbf{x}}$ remains faithful to the input geometry $\mathbf{x}$;
    \item \textbf{Codebook loss} $\Vert \mathrm{sg}[E_{\phi}(\mathbf{x})] - \mathbf{z}_q \Vert_2^2$, which updates the codebook embeddings toward the encoder outputs;
    \item \textbf{Commitment loss} $\beta \Vert E_{\phi}(\mathbf{x}) - \mathrm{sg}[\mathbf{z}_q] \Vert_2^2$, which penalizes large deviations between the encoder outputs and the assigned codewords, encouraging the encoder to commit to discrete representations. 
\end{itemize}
Here, $\beta$ is a weighting hyperparameter, typically set to $0.25$.

\vspace{0.3em}
\noindent\textbf{Optimization Strategy.}
In practice, the codebook can be updated either through the explicit codebook loss in Equation~\ref{eq:vqvae_loss} or through an exponential moving average (EMA), which improves stability by smoothing embedding updates over training iterations. However, both approaches mainly optimize the codewords that are actively selected during quantization, while rarely used codewords receive little training unless reactivated.

In our training method, the codebook is updated not only through standard gradient based updates of active codewords, but also through a targeted dynamic reinitialization mechanism inspired by CVQ-VAE~\cite{CVQ} that reactivates persistently unused entries. Throughout training, the system monitors the assignment statistics of all codewords and reinitializes those with consistently low utilization by drawing representative feature vectors from a historical feature buffer and directly replacing the corresponding inactive entries. In addition, we replace the conventional Euclidean distance nearest neighbor search in Equation~\ref{eq:vq} with cosine similarity based matching, which improves the generalization capability of the learned geometric representations.

\vspace{0.3em}
\noindent\textbf{Interpretation.}
The VQ-VAE effectively transforms high-dimensional continuous data into a discrete index representation, forming a finite vocabulary of latent tokens. This discrete latent space provides a symbolic abstraction of geometry, enabling autoregressive models, such as Transformer, to model geometric data analogously to natural language. Each latent index can thus be viewed as a \textit{Geometry Token} capturing local shape features, serving as the fundamental unit for downstream sequence modeling.

\subsection{Ablation Study of Face-Edge VQ-VAE}
In this section, we conduct an ablation study to analyze how the size of the VQ-VAE codebook influences reconstruction performance, in terms of both geometric fidelity and B-rep validity. VQ-VAE discretizes continuous encoder features into a latent space composed of $K$ learnable codebook vectors, and the choice of $K$ determines the representational capacity of this space, thereby affecting reconstruction quality and training stability. To isolate the effect of codebook size, we experiment with four configurations, specifically $K = 512$, $1,024$, $2,048$, and $4,096$, while keeping all other hyperparameters fixed. Experiments are performed on the DeepCAD dataset. Model performance under each setting is assessed using two metrics:
\begin{itemize}
    \item \textbf{Valid}: For each codebook size, we randomly sample 3,000 B-rep instances from the test set and use the same subset across all configurations. Given the predicted UV-sampled features together with the associated ground-truth geometric information (e.g., 3D positions), we apply the post-processing pipeline to reconstruct B-rep models and compute the proportion of reconstructions deemed valid.

    \item \textbf{Reconstruction Loss}: The geometry reconstruction loss on the training and validation sets (i.e., Train Loss and Val Loss), which measures the fidelity of the reconstructed shapes with respect to the input.
\end{itemize}

As shown in Table~\ref{tab:ablation_codebook}, increasing the codebook size consistently improves reconstruction fidelity and reduces the B-rep invalid rate. When the codebook size reaches $K = 4,096 $, the model obtains the best overall performance, exhibiting the highest valid rate and stable training dynamics without signs of overfitting. Although larger codebooks may theoretically provide higher representational capacity, they are prone to poor utilization and unstable quantization behavior during VQ-VAE training. Moreover, excessively large codebooks expand the discrete vocabulary used for downstream autoregressive modeling, which can hinder sample quality and introduce additional computational cost. Therefore, we adopt $K = 4,096 $ for the DeepCAD dataset as a balanced configuration that achieves strong reconstruction accuracy, high B-rep validity, stable quantization behavior, and manageable complexity for generative modeling.

\begin{table}[t]
\centering
\caption{Ablation study of codebook size in VQ-VAE.}
\label{tab:ablation_codebook}
\small
\setlength{\tabcolsep}{6pt}
\begin{tabular}{lccc}
\toprule
\textbf{Codebook Size} & 
\textbf{Valid~$\uparrow$} & 
\textbf{Train Loss~$\downarrow$} & 
\textbf{Val Loss~$\downarrow$} \\
\midrule
512  & 69.8\% & $9.2 \times 10^{-5}$ & $2.0 \times 10^{-5}$ \\
1024 & 73.2\% & $7.0 \times 10^{-5}$ & $1.0 \times 10^{-5}$ \\
2048 & 76.7\% & $4.7 \times 10^{-5}$ & $9.0 \times 10^{-6}$ \\
4096 & 87.5\% & $1.6 \times 10^{-5}$ & $3.0 \times 10^{-6}$ \\
\bottomrule
\end{tabular}
\vspace{-3pt}
\end{table}

\subsection{Ablation Design of 3D Position Tokenization}

This section investigates several deep learning architectures for vector quantization of bounding boxes, where each bounding box is represented as a six-dimensional vector $\mathbf{b} = [x_{\min}, y_{\min}, z_{\min}, x_{\max}, y_{\max}, z_{\max}] \in [-1,1]^6$. To evaluate and compare the accuracy of different quantization strategies for bounding boxes, we structure our ablation studies around two paradigms: \textit{convolution-based spatial expansion} schemes and \textit{Transformer-based quantization} schemes.

In the \textit{convolution-based spatial expansion} paradigm, to leverage the powerful convolutional architecture of VQ-VAE and adapt its 2D convolutions, we repeat the bounding box vector across spatial grids of varying resolutions and evaluate model performance under two configurations:

\begin{itemize}
    \item \textbf{Scheme 1 ($32\times32$ High-Res)}: We reshape $\mathbf{b}$ into a $(2,3)$ matrix, then repeat it to $(32,3)$, and finally tile it into $\mathbf{b}' \in \mathbb{R}^{32\times32\times3}$. We then feed $\mathbf{b}'$ into a five-block U-Net encoder with channel widths $32, 64, 128, 256, 512$, where the first four blocks perform downsampling and the last block extracts features without further downsampling, yielding a latent feature map $\mathbf{z}_e \in \mathbb{R}^{2\times2\times128}$.
    
    \item \textbf{Scheme 2 ($16\times16$ Low-Res)}: Similarly, we construct a lower-resolution input $\mathbf{b}'' \in \mathbb{R}^{16\times16\times3}$ and employ a lighter four-block U-Net backbone with channel widths $64, 128, 256, 512$, also yielding a latent feature map $\mathbf{z}_e \in \mathbb{R}^{2\times2\times128}$.
\end{itemize}

For both convolution-based schemes, we apply a $1\times1$ convolution to project $\mathbf{z}_e$ to 64 dimensions to match the codebook, perform nearest-neighbor lookup against a learnable codebook of size $4,096$ and dimensionality $64$ to obtain the quantized representation $\mathbf{z}_q$, and reconstruct the input via a symmetric decoder.

In the \textit{Transformer-based quantization} schemes, we directly perform feature extraction and vector quantization on the bounding box representation.

\begin{itemize}
    \item \textbf{Scheme 3 (Single-code quantization).}
    The entire bounding box is treated as a single atomic unit. A multilayer perceptron (MLP) maps $\mathbf{b}$ to a 64-dimensional embedding vector, which is then processed by an 8-layer standard Transformer encoder. The resulting feature is quantized into a single discrete codeword selected from the codebook. The decoder consists of a symmetric stack of Transformer encoder blocks, followed by an MLP that reconstructs the bounding box.

    \item \textbf{Scheme 4 (Multi-code quantization).}
    To explicitly capture both the independence and the inter-axis correlations among the $X$, $Y$, and $Z$ dimensions, we reshape $\mathbf{b}$ into a $(3,2)$ matrix, corresponding to the minimum and maximum values along each axis. An MLP embeds this matrix into a tensor of shape $(3,64)$, which is subsequently processed by the same 8-layer Transformer encoder architecture. Each of the three 64-dimensional features is independently quantized, producing three discrete codewords from the shared codebook. The decoder adopts a symmetric architecture mirroring that of the encoder, comprising a symmetric stack of Transformer encoder blocks followed by an MLP to reconstruct the original bounding box.

\end{itemize}
All Transformer-based schemes use an embedding dimension of 64, a feed-forward hidden dimension of 256, and a codebook of size 4,096 with 64-dimensional codewords. 

We evaluate the performance of the four schemes under two metrics:

\begin{itemize}
    \item \textbf{Acc}: Bounding box reconstruction accuracy, defined as the fraction of samples for which all six components of the reconstructed bounding box $\hat{\mathbf{b}}$ lie within an absolute tolerance $\epsilon$ of the ground-truth $\mathbf{b} \in [-1,1]^6$. Formally,
\begin{equation}
\begin{aligned}
    \mathrm{Acc} &= \frac{1}{|\mathcal{D}|} \sum_{\mathbf{b} \in \mathcal{D}} s(\mathbf{b}, \hat{\mathbf{b}}), \\
    s(\mathbf{b}, \hat{\mathbf{b}}) &=
    \begin{cases}
        1, & \text{if } |\hat{b}_i - b_i| \leq \epsilon \quad \forall i \in [6], \\
        0, & \text{otherwise}.
    \end{cases}
    \end{aligned}
    \end{equation}
    We set $\epsilon = 0.01$, with $|\mathcal{D}|$ denoting the size of the dataset.

    \item \textbf{Reconstruction Loss}: The geometry reconstruction loss on the training and validation sets (i.e., Train Loss and Val Loss), measured as the $\ell_2$ distance between the input bounding box and its VQ-VAE reconstruction. This metric quantifies the overall fidelity of shape reconstruction.
\end{itemize}

After extensive hyperparameter tuning, we report the performance of the four position tokenization schemes on the two metrics described above. The results are presented in Table~\ref{tab:position_token_ablation}. For deep learning-based methods, the best reconstruction accuracy under $\epsilon = 0.01$ reaches $99.4\%$. In comparison, our proposed \textit{uniform scalar quantization} partitions the interval $[-1,1]$ into $L$ discrete levels with a fixed step size of $2/(L - 1)$. Accordingly, the maximum reconstruction error is theoretically bounded by half of this step, i.e., $1/(L - 1)$. Therefore, setting $L = 101$ guarantees $100\%$ reconstruction accuracy when $\epsilon = 0.01$. Moreover, as $L$ increases, the \textit{uniform scalar quantization} scheme can achieve reconstruction accuracy that surpasses deep learning-based approaches, highlighting the superior precision and reliability of our method. Balancing precision and vocabulary size, we choose $L = 2,048$ for our final setting.

\begin{table}[t]
\centering
\caption{Ablation study of 3D position tokenization schemes.}
\small
\setlength{\tabcolsep}{10pt} 
\begin{tabular}{lccc}
\toprule
\textbf{Method} & 
\textbf{Acc~$\uparrow$} & 
\textbf{Train Loss~$\downarrow$} & 
\textbf{Val Loss~$\downarrow$} \\
\midrule
Scheme 1 & 98.4\% & $2.5 \times 10^{-5}$ & $1.5 \times 10^{-5}$ \\
Scheme 2 & 99.4\% & $2.0 \times 10^{-5}$ & $8.0 \times 10^{-6}$ \\
Scheme 3 & 96.4\% & $4.6 \times 10^{-5}$ & $3.3 \times 10^{-5}$ \\
Scheme 4 & 97.6\% & $3.9 \times 10^{-5}$ & $2.5 \times 10^{-5}$ \\
\bottomrule
\end{tabular}
\label{tab:position_token_ablation}
\vspace{-3pt}
\end{table}

\subsection{Qualitative Analysis of Block Sequentialization}
In our holistic token sequence representation, each geometric primitive (i.e., face or edge) is represented by a fixed-length token block. The full sequence is constructed by concatenating these blocks in a specific order. To better preserve causal ordering and maintain the inherent local geometric structure of the B-rep model during autoregressive generation, we explored different strategies for ordering these blocks. Specifically, our strategy involves two key objectives: 1) Face block ordering: Faces sharing an edge should be positioned close to each other in the sequence, and 2) Edge block ordering: Each edge should be positioned close to its associated face in the sequence. Thus, we aim to minimize the block-level distances between topologically related primitives. 

We denote the sets of face blocks and edge blocks by $\mathcal{F} = \{f_1, \dots, f_{N_f}\}$ and $\mathcal{E} = \{e_1, \dots, e_{N_e}\}$, respectively. For each face block $f$ and edge block $e$, let $p_F(f) \in \mathbb{N}$ and $p_E(e) \in \mathbb{N}$ represent the starting token positions of their corresponding blocks. The distance between any two blocks is defined as the absolute difference between their starting positions, representing their relative separation within the causal context window. Based on this notion of distance, we formulate the following two objectives:

\begin{itemize}
    \item \textbf{Face block ordering loss}:
    \begin{equation}
        L_F = \sum_{(f_i, f_j) \in \mathcal{A}_F} \big| p_F(f_i) - p_F(f_j) \big|
        \label{eq:L_F}
    \end{equation}
    where $\mathcal{A}_F = \{(f_i, f_j) \mid f_i \text{ and } f_j \text{ share an edge}\}$. 
    This loss measures the total sequential distance between all pairs of adjacent faces.
    
    \item \textbf{Edge block ordering loss}:
    \begin{equation}
        L_E = \sum_{e \in \mathcal{E}} \sum_{f \in \mathcal{F}(e)} \big| p_E(e) - p_F(f) \big|
        \label{eq:L_E}
    \end{equation}
    where $\mathcal{E}$ is the set of all edge blocks, and $\mathcal{F}(e)$ denotes the set of face blocks incident to edge block $e$. This loss measures the total sequential distance between each edge block and the blocks of all its incident faces.

\end{itemize}
\noindent\textbf{Analysis of face block ordering loss.} 
Minimizing the Face block ordering loss $L_F$ (Equation~\ref{eq:L_F}) corresponds to the classical \textit{Minimum Linear Arrangement} (MLA) problem on the face adjacency graph, which seeks a linear ordering of nodes that minimizes the total distance between adjacent pairs. However, standard MLA formulations assume symmetric context access and do not account for the unidirectional dependency inherent in autoregressive generation. In a decoder-only architecture with causal masking, a face block can attend only to previously generated blocks. This results in an asymmetric information flow: a face generated later can condition on its earlier neighbors, but not vice versa. Under this constraint, the conventional MLA objective is no longer fully appropriate. To accommodate this causal structure, we introduce a structural constraint: faces should be ordered in \textit{descending degree} (i.e., faces with higher degree are placed earlier in the sequence). This ensures that faces with richer connectivity appear early, thereby maximizing the opportunity for their adjacent faces to attend to them during subsequent generation steps. Consequently, the face block ordering problem becomes a \textit{Minimum Linear Arrangement problem with a descending-degree constraint}, preserving the core MLA objective while explicitly incorporating the causal prior required by autoregressive modeling.

\noindent\textbf{Analysis of edge block ordering loss.} 
In our sequence layout, all face blocks appear before a single \textit{SEP} token, followed by all edge blocks. Each face block consists of a fixed number of $k_f$ tokens, and each edge block consists of a fixed number of $k_e$ tokens. Let $p_{\text{SEP}}$ denote the token position of the \textit{SEP} token.

For any edge block $e$ and an incident face block $f$, the block-level distance between their starting positions decomposes as:
\begin{equation}
L_E = 
\underbrace{\sum_{e \in \mathcal{E}} \sum_{f \in \mathcal{F}(e)} \!\! \bigl(p_E(e) - p_{\text{SEP}}\bigr)}_{\text{edge-to-SEP}} 
+ 
\underbrace{\sum_{e \in \mathcal{E}} \sum_{f \in \mathcal{F}(e)} \!\! \bigl(p_{\text{SEP}} - p_F(f)\bigr)}_{\text{face-to-SEP}}
\end{equation}
Here, the total block-level distance between incident face blocks $f$ and edge blocks $e$ decomposes into two aggregated components: the \textit{face-to-SEP} term, which quantifies the total distance of the face blocks to the separator, and the \textit{edge-to-SEP} term, which quantifies the total distance of the edge blocks to the separator.

We observe that the edge-to-SEP distances form an arithmetic sequence, since all edge blocks are placed contiguously right after the \textit{SEP} token. Consequently, the sum of these distances is:
\begin{equation}
\text{edge-to-SEP} = N_e + k_e \cdot \frac{(N_e - 1)N_e}{2}
\end{equation}
Here, \(N_e\) denotes the number of edge blocks and \(k_e\) is the size (i.e., number of tokens) of each edge block, both of which are fixed constants for a given input. The first edge block is at distance 1 from \textit{SEP}, and each subsequent block is shifted by \(k_e\) positions. Since this sum depends only on the constants \(N_e\) and \(k_e\), the edge-to-SEP contribution to the edge block ordering loss \(L_E\) (Equation~\ref{eq:L_E}) remains unchanged under any permutation of the edge blocks. Consequently, \(L_E\) is determined solely by the ordering of the face blocks.

For face-to-SEP, we observe that a face block \( f \) of degree \( \deg(f) \) contributes its distance to \textit{SEP} exactly \( \deg(f) \) times, once for each incident edge. This yields the expression:
\begin{equation}
\text{face-to-SEP} = \sum_{f \in \mathcal{F}} \deg(f) \cdot (p_{\text{SEP}} - p_F(f))
\label{eq:L_E_simplified}
\end{equation}
where \( \mathcal{F} \) denotes the set of all face blocks. Moreover, when generating any edge block, the full set of faces is already available as context, so the causal dependency for edge prediction is naturally satisfied. Therefore, to reduce \( L_E \), faces with higher degree should be placed closer to \textit{SEP}, which means positioning them later in the face block sequence to effectively minimize the weighted sum of their distances.

\noindent\textbf{Balanced Ordering Strategies.} 
Based on our analyses of the Face and Edge block ordering losses, we find that the two objectives impose mutually conflicting constraints on where high-degree faces should be placed in the sequence:

\begin{itemize}
    \item Minimizing \(L_F\) corresponds to a minimum linear arrangement (MLA) problem with a \textit{high-degree-first} constraint: high-degree faces should be placed early in the sequence.
    \item Minimizing \(L_E\) requires placing high-degree faces late in the sequence, near \textit{SEP}, to minimize their weighted distance contribution.
\end{itemize}

This leads to a multi-objective constrained minimum linear arrangement problem, for which we propose the following heuristic strategies:

\begin{itemize}
    \item \textbf{Prioritize minimizing \(L_F\)}:  
    \begin{itemize}
        \item \textit{Spectral Ordering (SS)}:  
        Perform spectral ordering of faces using the Fiedler vector, which is the second smallest eigenvector of the graph Laplacian and a standard technique for approximating solutions to the MLA problem. After obtaining the initial ordering, compute the average degree of faces in the first and second halves of the sequence. If the second half exhibits a higher average degree, reverse the entire sequence to place high-degree faces earlier.
    \end{itemize}

    \item \textbf{Prioritize minimizing \(L_E\)}:  
    \begin{itemize}
        \item \textit{Degree-Ascending Ordering (DEG-A)}:  
        Sort faces by degree in ascending order so that low-degree faces come first and high-degree faces appear last. This ordering strictly minimizes the edge--face distance term in \(L_E\).
    \end{itemize}

    \item \textbf{Balanced strategies} that jointly account for both the Face block ordering loss (\(L_F\)) and Edge block ordering loss (\(L_E\)):
    \begin{itemize}
        \item \textit{Centroid-based Ordering (ZYX)}: Sort faces lexicographically by their centroid coordinates in the order Z, then Y, then X;
        \item \textit{Breadth-First Search (BFS)}:  
        Start from the highest-degree face and perform a BFS that preferentially expands toward lower-degree neighboring faces.
        \item \textit{Depth-First Search (DFS)}:  
        Start from the highest-degree face and similarly prioritize visiting lower-degree neighbors during traversal.
    \end{itemize}
\end{itemize}

The ordering of edge blocks does not affect the values of $L_F$ or $L_E$. However, to avoid excessively large differences between $p_E(e)$ and $p_F(f)$ that could destabilize the attention mechanism in autoregressive modeling, we adopt the \textit{maximum adjacent face index ascending} strategy (MAX-IDX-A). Specifically, for each edge, we define its sorting key as the maximum index among all its adjacent faces in the face sequence, and then sort all edges in ascending order of this key. This approach encourages each edge to appear close to its neighboring faces in the sequence, thereby improving the distribution of edge--face positional offsets.
The experimental results for the face and edge block ordering strategies can be found in Section~\ref{Ablation Studies}.